\documentclass{article}

\usepackage[english]{babel}

\usepackage[letterpaper,top=2cm,bottom=2cm,left=3cm,right=3cm,marginparwidth=1.75cm]{geometry}

\usepackage{amsmath}
\usepackage{graphicx}
\usepackage[colorlinks=true, allcolors=blue]{hyperref}
\usepackage{multirow}

\usepackage{authblk}
\usepackage{setspace}
\usepackage{subcaption}
\usepackage{amsmath}
\usepackage{amssymb}
\usepackage[numbers, compress]{natbib}
\usepackage{physics}

\usepackage{authblk}
\usepackage{hyperref}
\usepackage{caption} 

\title{Quantum Support Vector Machine for Prostate Cancer Detection: A Performance Analysis}

\author[1]{W. El Maouaki\thanks{Corresponding author. Email: \href{mailto:email@example.com}{walid.elmaouaki-etu@etu.univh2c.ma}}}

\author[1]{T. SAID}
\author[1,2]{M. BENNAI}

\affil[1]{Quantum Physics and Magnetism Team, LPMC, Faculty of Sciences Ben M’Sik, Hassan II University of Casablanca, Morocco}
\affil[2]{Lab of High Energy Physics, Modeling and Simulations, Faculty of Sciences, University Mohammed V-Agdal, Rabat, Morocco}

\date{}
\begin{document}
\maketitle
\begin{abstract}
This study addresses the urgent need for improved prostate cancer detection methods by harnessing the power of advanced technological solutions. We introduce the application of Quantum Support Vector Machine (QSVM) to this critical healthcare challenge, showcasing an enhancement in diagnostic performance over the classical Support Vector Machine (SVM) approach. Our study not only outlines the remarkable improvements in diagnostic performance made by QSVM over the classic SVM technique, but it delves into the advancements brought about by the quantum feature map architecture, which has been carefully identified and evaluated, ensuring it aligns seamlessly with the unique characteristics of our prostate cancer dataset. This architecture succeded in creating a distinct feature space, enabling the detection of complex, non-linear patterns in the data. The findings reveal not only a comparable accuracy with classical SVM ($92\%$) but also a $7.14\%$ increase in sensitivity and a notably high F1-Score ($93.33\%$). This study's important combination of quantum computing in medical diagnostics marks a pivotal step forward in cancer detection, offering promising implications for the future of healthcare technology.

\vspace{3mm}


\end{abstract}

\section{Introduction}

Prostate cancer, characterized by its high incidence rate among men worldwide \cite{rawla2019epidemiology}, presents a critical healthcare challenge \cite{culp2020recent}. The prognosis of this disease is heavily dependent on the timeliness and accuracy of its detection. Early-stage identification of this cancer not only offers a greater array of therapeutic options but also significantly enhances the chances of successful treatment outcomes \cite{olabanjo2023application, cuocolo2019machine}.  This underscores the urgent need for advanced diagnostic methodologies that can effectively intercept the disease in its nascent stages, thereby altering its path toward a more positive prognosis.

Machine learning, particularly through the implementation of support vector machines (SVMs), has marked a significant step forward in the interpretation of intricate biomedical data, which could be pivotal in addressing the early detection challenges of prostate cancer \cite{akinnuwesi2023application}. While traditional SVMs have demonstrated proficiency in data classification \cite{sweilam2010support, gaye2021improvement, dubey2023advancing, li2018support}, they face significant constraints in scalability and computational efficiency when applied to the expansive and high-dimensional datasets typical in prostate cancer research. This limitation calls for innovative approaches that can effectively harness the full potential of machine learning in this critical field of healthcare.

In the quest to overcome the limitations imposed by traditional SVMs in the landscape of oncological diagnostics, Quantum Support Vector Machines (QSVMs) have emerged as a frontier technology \cite{rebentrost2014quantum, li2015experimental}. This algorithm leverages the principles of quantum computing, such as superposition and entanglement, to process information in a fundamentally different manner. Unlike their classical counterparts, QSVMs are uniquely equipped to handle multidimensional data. This is due to their ability to perform complex linear algebra computations, which are the core of SVMs, exponentially faster on quantum processors. QSVMs offer a promising avenue to not only match but potentially surpass the accuracy of classical SVMs, with the added benefit of reduced computational times \cite{tychola2023quantum, gentinetta2022complexity}.

This study is based on the systematic evaluation of Quantum Support Vector Machines (QSVMs) to determine their effectiveness in the intricate field of prostate cancer detection. A pivotal aspect of this work is the identification and evaluation of an optimal feature map architecture, specifically tailored to the dataset at hand, with its mathematical framework meticulously detailed. We aim to rigorously benchmark the QSVM's diagnostic performance against traditional SVMs across all performance metrics, encompassing accuracy, specificity, sensitivity, and the F1 score, thereby quantifying the practical advantages of quantum-enhanced machine learning. Additionally, the research is designed to rigorously assess the QSVM's operational feasibility on actual clinical data, exploring its robustness and diagnostic ability. The comparative study highlights that our QSVM matches the accuracy of traditional SVMs and surpasses them in sensitivity and F1 score due to the high dimensional advantage of quantum Hilbert space. This underlines QSVM's improved detection of true positives, crucial for clinical reliability and reducing false negatives in prostate cancer screening.

\section{Methodology}
To evaluate the potential benefits of QSVM in prostate cancer detection, a comparative analysis was conducted using a dataset derived from prostate imaging results and patient attributes. The data were preprocessed and transformed into a quantum-ready format suitable for QSVM classification. The performance of QSVM was compared to classical SVM using performance metrics (accuracy, precision, sensitivity, specificity, and F1-Score).

\subsection{Dataset and Processing Steps}
Quantum information processing systems are characterized by features like quantum superposition and quantum entanglement, giving them the potential to enhance image recognition speeds in machine learning. Certain quantum image processing techniques have demonstrated the potential to outpace their classical equivalents dramatically \cite{9362910, PhysRevX.7.031041, su2021improved}. Quantum classifiers have been applied to diverse image datasets in the literature \cite{10126421, maheshwari2021variational, JADHAV20232612, wu2023quantum, wei2023quantum}. In this study, our attention is on the Prostate cancer dataset, which is widely perceived as a significant challenge.
% Notably, numerous conventional supervised learning methods have achieved nearly flawless outcomes for this dataset [114]. 
The dataset used in this study is the Kaggle Prostate Cancer Dataset \cite{sajid_prostate_2023}. The dataset consists of 100 observations and 9 variables (out of which 8 numeric variables and one categorical variable) which are as follows: Radius, Texture, Perimeter, Area, Smoothness, Compactness, Symmetry, Fractal dimension, diagnosis result (labels). This problem is a two-category classification problem: Cancerous and non-Cancerous Prostate.

\smallskip

In the data preprocessing phase, our initial dataset exhibited an imbalance, with a predominance of label $1$ over label $0$ data points. This can potentially cause machine learning models to favor the dominant label, leading to the misclassification of the minority label. To counter this, we utilized the \textit{RandomOverSampler} class, strategically duplicating examples from the minority class to balance the training dataset. Subsequently, we standardized the dataset using the \textit{StandardScaler} class from the \textit{sklearn.preprocessing} module. This process scales the features to have a mean of 0 and a standard deviation of $1$, enhancing feature comparability and often being a prerequisite for optimal machine learning performance. These operations extend our data to 124 data samples. For QSVM, we added a normalization step using the \textit{MinMaxScaler} class. This scaled the features to range between $0$ and $1$, aiding model interpretability and facilitate the subsequent quantum data encoding process, as detailed in Section \ref{QSVM_method}. In this encoding process, features are mapped onto rotation gates, operations that are optimized for values (or angles) in the $0$ to $1$ range. The conversion function of the MinMax method \cite{jin2015data} is given by equation \ref{MinMax}. Finally, the processed data was divided into training ($80\%$) and testing ($20\%$) subsets.

\begin{equation}
    X_S=\frac{X-X \cdot \min (\text { axis }=0)}{X \cdot \max (\operatorname{axis}=0)-X \cdot \min (\operatorname{axis}=0)}(\max -\min )+\min
    \label{MinMax}
\end{equation}

\smallskip

SVM and QSVM can be sensitive to the scale and distribution of the input data. Hence, preprocessing is crucial for achieving meaningful and accurate results. In the following section, we'll discuss how SVM and QSVM work and how they were employed to classify our prepared data.

\subsection{Support Vector Machines (SVMs)}

\begin{figure}[ht]
    \centering
    \includegraphics[width=0.5\linewidth]{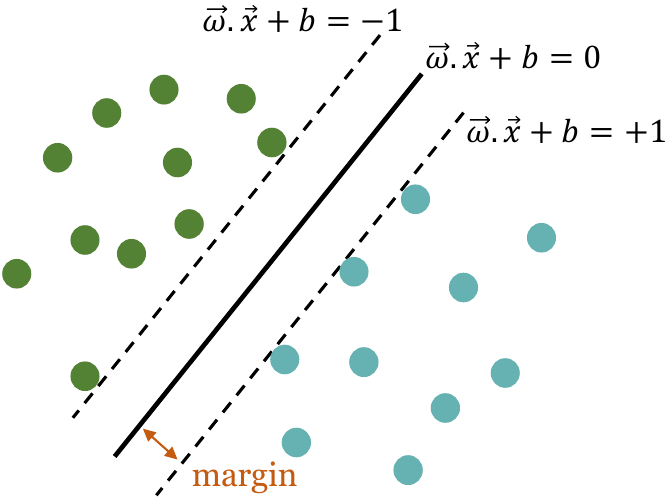}
    \caption{ Illustration of the maximum-margin hyperplane for SVM classification. The margin hyperplanes for support vectors are depicted by the dashed lines.}
    \label{hyperplane_svm}
\end{figure}

SVMs are a pivotal technique in machine learning, primarily designed for binary classification tasks. At their core, SVMs work by integrating linear algebra and optimization. Given data vectors, denoted as $\mathbf{x} \in \mathbb{R}^r$ with $r \geq 1$, each associated with a class label -1 or 1, the primary objective of SVMs is to identify the optimal hyperplane that separates these classes with the broadest margin. SVM constructs a hyperplane described by the equation $\vec{w} \cdot \vec{x}+b=0$. For a given training instance $\vec{x}_i$ belonging to the positive category, the condition $\vec{w} \cdot \vec{x}+b \geq 1$ must hold. Conversely, for $\vec{x}_i$ in the negative category, the condition $\vec{w} \cdot \vec{x}+b \leq-1$ is expected, see Fig. \ref{hyperplane_svm}.

More formally, we extract a set of training samples, $X=\left\{x_1, \ldots, x_M\right\}$, from an underlying probability distribution $P(x, y)$, and each vector has an associated label $y=\{y_1, \ldots, y_M\}$. Using this data, SVMs develop a function, $f: \mathbb{R}^r \rightarrow \{-1,1\}$. This function is tasked with two primary goals: achieving high accuracy in predicting the actual labels and maximizing the margin, which is the distance between the two classes on the hyperplane. Once the SVM is trained with this function, it is equipped to classify new data points that are sampled from the same probability distribution.

The decision function for a given new input $x$ is:
\begin{equation}
f(x)=\operatorname{sign}\left(w^T x+b\right)
\end{equation}

Where $w$ represents the weight vector, $x$ denotes the feature vectors, and $b$ is the bias term. To achieve this optimal hyperplane, SVM employs a convex optimization strategy that minimizes the following \textbf{primal optimization problem}:

\begin{equation}
    \min _w \frac{1}{2}\|w\|^2+C \sum_{i=1}^n \max \left(0,1-y_i\left(w^T x_i+b\right)\right)
    \label{primal_prob}
\end{equation}

Here, the term $\max \left(0,1-y_i\left(w^T x_i+b\right)\right)$ represents the hinge loss, which measures the misclassification or violation of the margin. $C$ is a regularization parameter, balancing the trade-off between maximizing the margin and minimizing the contribution of the hinge loss.

We begin by presenting the standard kernelized support vector machines from classical machine learning. We then describe how these techniques can be extended to the quantum setting by using quantum circuits to define the feature maps. To wrap up this section, we show the variational quantum circuits that compute the kernel for our application case.

\subsubsection{Kernelized support vector machines}
A key advantage of SVMs is their ability to efficiently perform nonlinear classification using kernel functions, which implicitly map the input data into high-dimensional feature spaces \cite{6524743}. Formally, let's represent this mapping as $\phi(\vec{x})$, where $\phi$ denotes the feature map that transforms input vectors $\vec{x}$ into a higher dimensionality.

The effectiveness of SVMs in this context lies in the 'kernel trick', a technique that circumvents the need for explicit computation of these high-dimensional transformations. Instead, SVMs utilize the dot product between vectors in the transformed space. This is achieved by introducing a kernel function $K\left(\vec{x}_i, \vec{x}_j\right)$, which effectively represents the dot product in this elevated feature space:
\begin{equation}
    K\left(\vec{x}_i, \vec{x}_j\right)=\phi\left(\vec{x}_i\right) \cdot \phi\left(\vec{x}_j\right)
    \label{kernel_equation}
\end{equation}

One of the popular kernel functions is the Radial Basis Function (RBF) 
\begin{equation}
    K\left(\vec{x}_i, \vec{x}_j\right)=\exp \left(-\frac{\left\|\vec{x}_i-\vec{x}_j\right\|^2}{2 \sigma^2}\right)
\end{equation}

where $\sigma$ is the width parameter. The RBF kernel allows SVMs to effectively delineate non-linear boundaries without the need to directly engage with computationally intensive high-dimensional spaces.

Rather than addressing the primal optimization problem, contemporary methods \cite{platt1998sequential} typically focus on the \textbf{dual form of equation \ref{primal_prob}}:

\begin{equation}
\begin{aligned}
& \max _\alpha \sum_{i=1}^n \alpha_i-\frac{1}{2} \sum_{i, j=1}^n \alpha_i \alpha_j y_i y_j K\left(x_i, x_j\right) \\
& \text { subject to } 0 \leq \alpha_i \leq C \\
& \sum_{i=1}^n \alpha_i y_i=0
\end{aligned}
\end{equation}
Once the Lagrange multipliers $\alpha_i$ are computed, the decision function in the transformed space becomes:
\begin{equation}
f(x)=\operatorname{sign}\left(\sum_{i=1}^n \alpha_i y_i K\left(x_i, \vec x\right)+b\right)
\label{decision_function}
\end{equation}

Here, it's important to emphasize that only the support vectors (data points for which $\alpha_i>$ 0) actually affect this decision. This aspect highlights the efficiency of SVM, as it uses only a subset of the training data to make predictions, rather than the entire dataset. Additionally, we can represent the weight vector $w$ in terms of these multipliers and the training data as $w=\sum_{i=1}^n \alpha_i y_i x_i$.

\subsection{Quantum Support Vector Machine (QSVM)} \label{QSVM_method}
QSVM is an innovative fusion of quantum computing and SVM, initially introduced by Rebentrost et al. \cite{rebentrost2014quantum}, and has since evolved, with further variants and enhancements being developed \cite{jadhav2023quantum, lin2020quantum, innan2023enhancing}. The QSVM approach presents a unique method for data processing and analysis. In traditional SVM, the time complexity scales as $O(poly(NM))$, with $N$ representing the feature space dimensions and $M$ the number of training samples. In contrast, QSVMs demonstrate a performance of $O(log(NM))$ in both training and classification phases \cite{zhaokai2014experimental}. This significant acceleration is attributed to the use of quantum circuits, which enable parallel computation of vector inner products.

In the classical case, we have used the tractable kernel corresponding to radial basis functions. However, we suggest using a quantum feature map to compute this kernel, as in equation \ref{kernel_equation}, which should be assessed on a quantum computer. In the quantum case, we encode our data vector $x$ into a quantum state $|\psi(x)\rangle$, where $|\psi(.)\rangle$ is a feature map. The feature map could be expressed as an application of a $2^n \times 2^n$ unitary operator $U$ to the quantum state $|0\rangle^n$ as $|\phi(x)\rangle=U_{\phi(x)}| 0^n\rangle$ (the full expression is in equation \ref{U_gate}), n is the dimension of the data.

To adapt the function in equation \ref{decision_function} for a quantum machine, it's important to select an appropriate quantum feature map. This decision subsequently determines the design of our quantum circuit. 

Several methods are available for picking an apt feature map, as discussed in \cite{PhysRevLett.122.040504}, such feature maps we can name ZFeatureMap, PauliFeatureMap, and ZZFeatureMap who play a critical role in encoding classical data into quantum states. These feature maps employ quantum gates like Pauli rotations (X, Y, Z) to convert data features into rotations in the quantum state space. They crucially introduce entanglement between qubits using gates like controlled-Z, allowing the quantum system to represent complex data correlations in high-dimensional Hilbert spaces. These entanglements differ in topology patterns, for instance, some employ nearest-neighbor entanglements, creating connections between adjacent qubits, whereas others use an all-to-all (fully connected) entanglement scheme, linking every qubit with every other. This diversity in entanglement topologies allows for varying degrees of complexity in representing data relationships.

\begin{figure}[ht]
    \centering
    \includegraphics[width=1\linewidth]{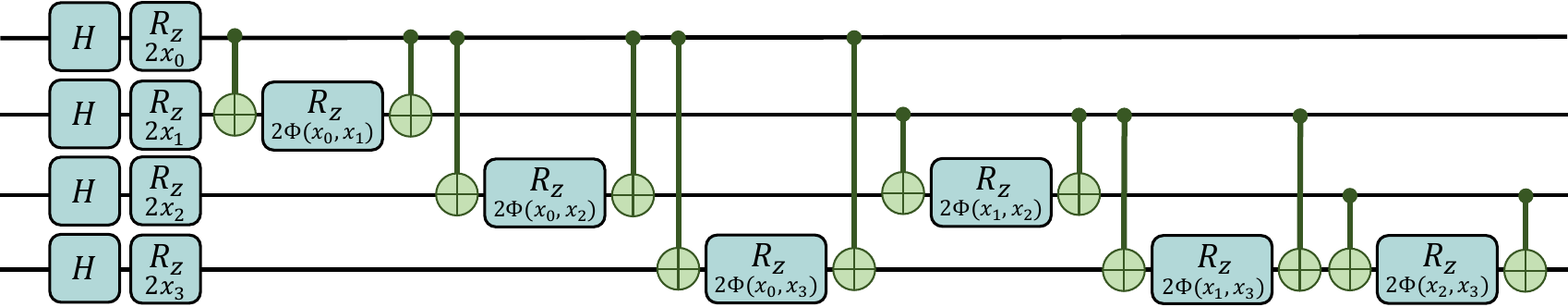}
    \caption{ZZ-Feature Map quantum circuit}
    \label{zzfeaturemap}
\end{figure}

In our case, we explored and determined the best feature map architecture for our dataset, and subsequently the ZZFeaturemap fully entanglement architecture succeeded in capturing the richness of our Prostate cancer dataset. For example, Fig. \ref{zzfeaturemap} depicts a $4$ qubits quantum circuit of the ZZFeaturemap fully entanglement feature map designed to process a dataset comprising $4$ dimensions.
Our Prostate feature data is 8 dimensional, so the feature map will apply 8 qubits ($2^8 \times 2^8$) unitary operator $U$ to obtain estimates of kernel $K$. The kernel function can be computed using the Hilbert-Schmidt inner product of the quantum states obtained from the two data points $x_i$ and $x_j$.

\begin{equation}
    K(x_i, x_j)=|\langle 0^n|U_{\phi(x_j)}^{\dagger} U_{\phi(x_i)}| 0^n\rangle|^2=|\langle\psi(x_i) \mid \psi(x_j)\rangle|^2
    \label{kernel_eq}
\end{equation}
Where $\langle\psi(x_i) \mid \psi(x_j)\rangle$ is the inner products (or the overlap) between feature representations (or quantum states) of two data points and can be estimated using a quantum circuit, as depicted in the Fig. \ref{QSVM_circuit}.

The ZZ feature mapping unitary operation is expressed in equation \ref{U_gate}, and implemented in Fig. \ref{zzfeaturemap}:

\begin{equation}
|\phi(\boldsymbol{x})\rangle=U_{\phi(\boldsymbol{x})}\left|0^{\otimes n}\right\rangle=U_{2^n}^{\mathrm{full-ent}}\left(\otimes_{q=1}^n\left(R_z\left(x_q\right) H\right)\right)\left|0^{\otimes n}\right\rangle
\label{U_gate}
\end{equation}
and
\begin{equation}
U^{\text{full-ent }}:=\prod_{q=1}^{n-1} \prod_{k=q+1}^n \mathbf{E}_{q, k}
\end{equation}
where
\begin{equation}\mathbf{E}_{q, k}=e^{-i \Phi(x_q, x_k) Z_q Z_k}, \;\;\; \Phi(x_q, x_k)=(\pi-x_q)(\pi-x_k)\end{equation}

$Z$ is the is the Pauli operator. Next, we explain how to compute the complete kernel matrix for n data points from the measurement of the quantum circuit.

\subsubsection{Derivation of the Kernel Matrix from Quantum Measurements}

To determine the kernel value in equation \ref{kernel_eq}, we create a quantum state by applying the quantum feature map of $x_i$ and then the adjoint of the quantum feature map for $x_j$ sequentially in the quantum circuit:
\begin{equation}
|\Psi\rangle=|\phi\left(x_i\right)\rangle \otimes |\phi^{*}\left(x_j\right)\rangle
\end{equation}

Here, $|\phi^{*}\left(x_j\right)\rangle$ represents the adjoint operation of the quantum feature map for $x_j$. After that, we measure the state $|\Psi\rangle$. The probability $P_0$ of observing the outcome $|0\rangle ^{\otimes n}$ on all qubits is:
\begin{equation}
    P_0=|\langle ^{n\otimes} 0 \mid \Psi\rangle|^2
\end{equation}

\begin{figure}[ht]
    \centering
    \includegraphics[width=0.8\linewidth]{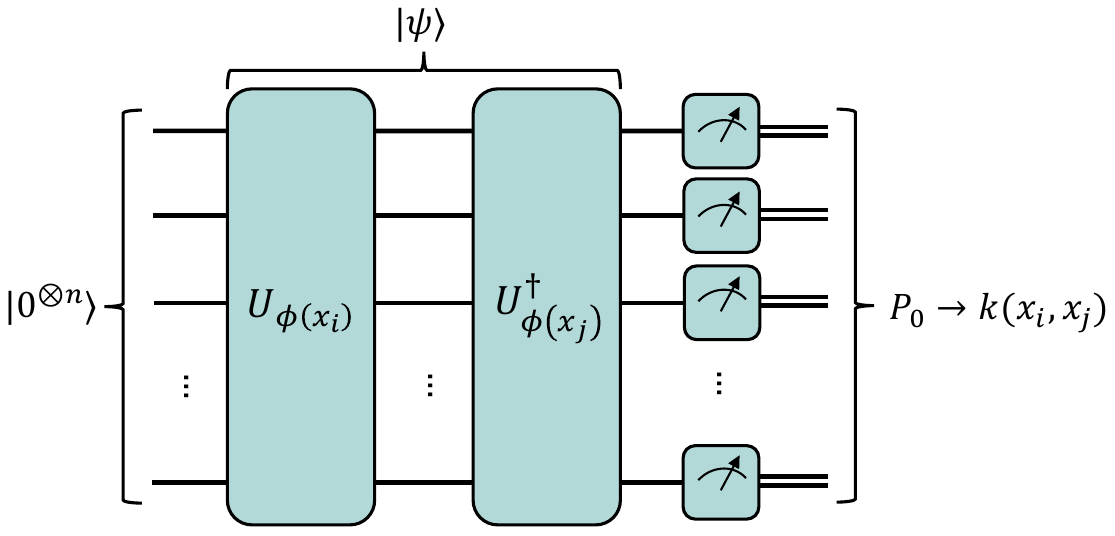}
    \caption{Quantum circuit to compute the kernel function, which can be approximated by assessing the occurrence frequency of $|0\rangle^{\otimes n}$ in the output.}
    \label{QSVM_circuit}
\end{figure}

This probability, when properly normalized, equates to the kernel value $K\left(x_i, x_j\right)$, see Fig. \ref{QSVM_circuit}. By iterating over all pairs of data points and extracting the corresponding $P_0$ values, one can construct the kernel matrix $K$. Each element $K_{i j}$ represents the kernel value between the data points $x_i$ and $x_j$. Subsequently, this quantum kernel matrix can be applied to various kernel-based algorithms, specifically, in our instance, to support vector machines as indicated in equation \ref{decision_function}. Utilizing quantum measurements in the QSVM process allows us to benefit from quantum parallelism and interference, making the computation of the kernel matrix more efficient than its classical counterpart.

In this study, we employed the quantum simulator framework offered by Qiskit, particularly utilizing the qasm simulator backend. 

\subsection{Advantages of QSVM}

QSVMs offer several benefits over classical SVMs:

\begin{enumerate}
    \item  Speed and Scalability: One of the most prominent advantages of QSVMs is their ability to perform computations much faster than classical SVMs. Quantum computing leverages quantum superposition and entanglement, enabling parallel processing of vast datasets. This is particularly crucial in cancer detection, where the analysis of large-scale genetic and molecular data is required. QSVMs can process these extensive datasets more efficiently, potentially reducing the time for pattern recognition and classification.

    \item Enhanced Feature Space Mapping: QSVMs have a natural advantage in mapping data into a high-dimensional feature space. Unlike classical SVMs, which require explicit kernel functions to map input data into higher-dimensional spaces, QSVMs can implicitly perform this mapping using quantum feature maps. This implicit mapping is more efficient and can handle more complex data structures, vital for identifying intricate patterns in cancer datasets.

    \item Quantum Kernel Estimation: QSVMs utilize quantum kernel estimation, which can lead to more accurate and nuanced separations in data classification. This quantum kernel trick can explore correlations and patterns in the data that classical kernels might miss or oversimplify. In cancer detection, this means potentially more accurate identification of malignant versus benign cells or more precise staging of cancer.

    \item Handling Noisy Data: Quantum algorithms, including QSVMs, are believed to be more robust against noise in data. This robustness is particularly valuable in medical datasets, which often contain uncertainties or incomplete information. QSVMs can provide more reliable analysis under these conditions.

    \item Quantum Data Loading: QSVMs can potentially benefit from quantum data loading techniques, which allow for the efficient input of large amounts of data into a quantum system \cite{cortese2018loading}. This capability is essential for dealing with high-dimensional data in cancer detection, where each sample can include thousands of genetic markers.
\end{enumerate}

% \subsection{Experimental Setups }

\subsection{Performance metrics}
In assessing the efficacy of the two classification methods in our study, we employed a suite of standard performance metrics. These metrics enable a comprehensive understanding of each method's strengths and weaknesses in various aspects of classification \cite{de2022general}. Below, we provide definitions and equations for each metric.

Accuracy is a metric that measures the overall correctness of the classification model. It is the proportion of true results (both true positives and true negatives) in the total number of instances.

$$
\text { Accuracy }=\frac{\text { True Positives (TP) }+\text { True Negatives (TN) }}{\text { Total Number of Instances}}
$$

Precision evaluates the model's exactness, indicating the proportion of positive identifications that were actually correct. It is particularly useful in scenarios where the cost of a false positive is high.
$$
\text { Precision }=\frac{\text { True Positives (TP) }}{\text { True Positives (TP) }+ \text { False Positives (FP) }}
$$

Sensitivity (also known as Recall) assesses the model's ability to correctly identify all relevant instances (true positives). It is crucial in contexts where missing any positive instance (such as a disease in medical diagnostics) is undesirable.

$$
\text { Sensitivity }=\frac{\text { True Positives (TP) }}{\text { True Positives (TP) }+ \text { False Negatives (FN) }}
$$

Specificity is a metric that measures the proportion of true negatives correctly identified. It is especially important in situations where avoiding false alarms is crucial.
$$
\text { Specificity }=\frac{\text { True Negatives (TN) }}{\text { True Negatives (TN)+False Positives (FP) }}
$$

The F1-Score is a harmonic mean of precision and sensitivity. It provides a balance between precision and sensitivity, offering a single metric to assess a model's accuracy when a class imbalance is present.
$$
\text { F1-Score }=2 \times \frac{\text { Precision } \times \text { Sensitivity }}{\text { Precision }+ \text { Sensitivity }}
$$

\section{Experimental results}

In this section, we present our experimental findings as obtained from the SVM and QSVM algorithms on the Prostate cancer dataset. The following figures elucidate the differences in how each model processes the dataset in its feature space and the subsequent impact on classification performance.

\smallskip

\begin{figure}[ht]
    \centering
    \includegraphics[width=0.6\linewidth]{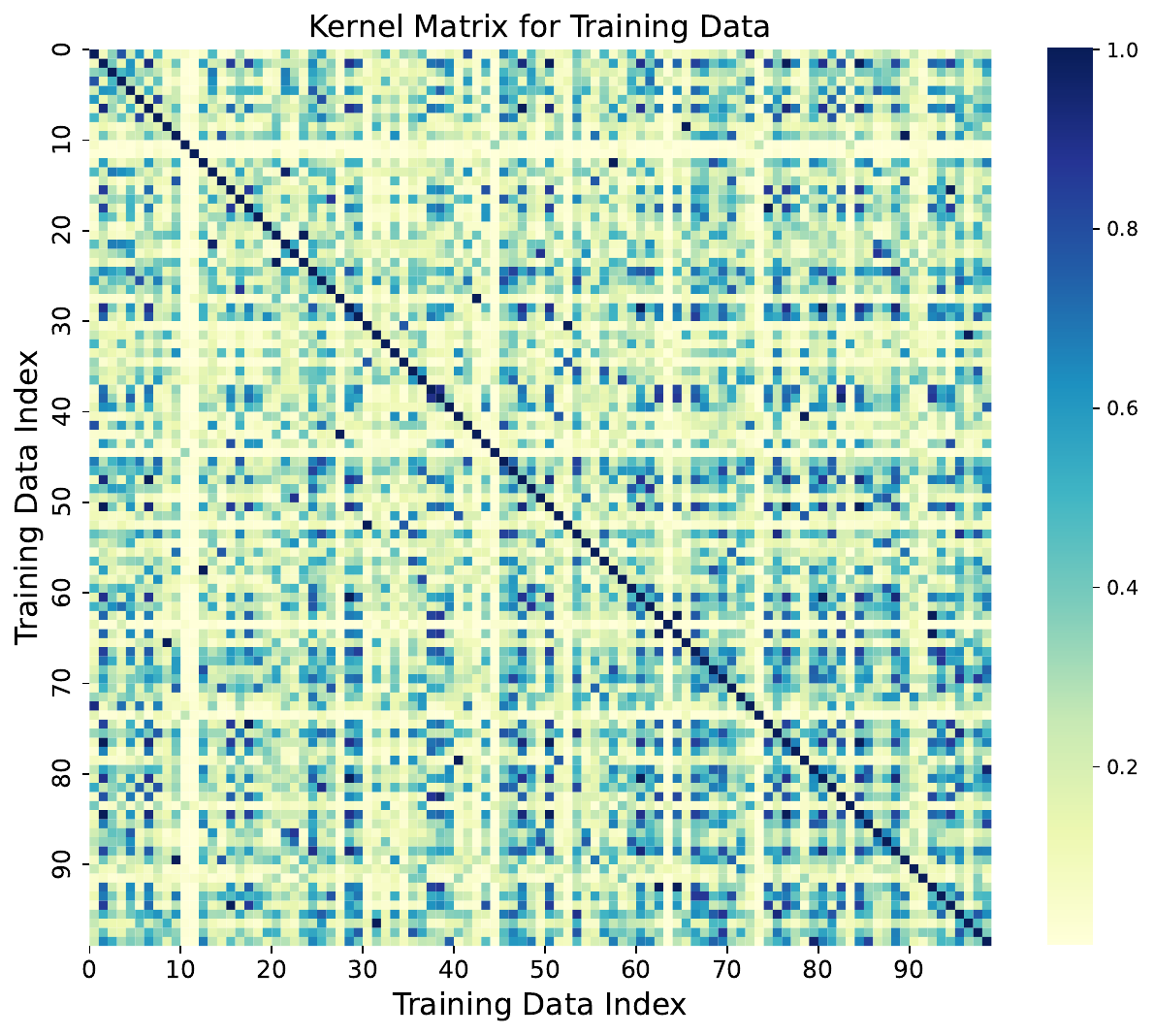}
    \caption{Kernel matrix for SVM training data using RBF kernel}
    \label{figure:SVM_Kernel}
\end{figure}

\begin{figure}[ht]
    \centering
    \includegraphics[width=0.6\linewidth]{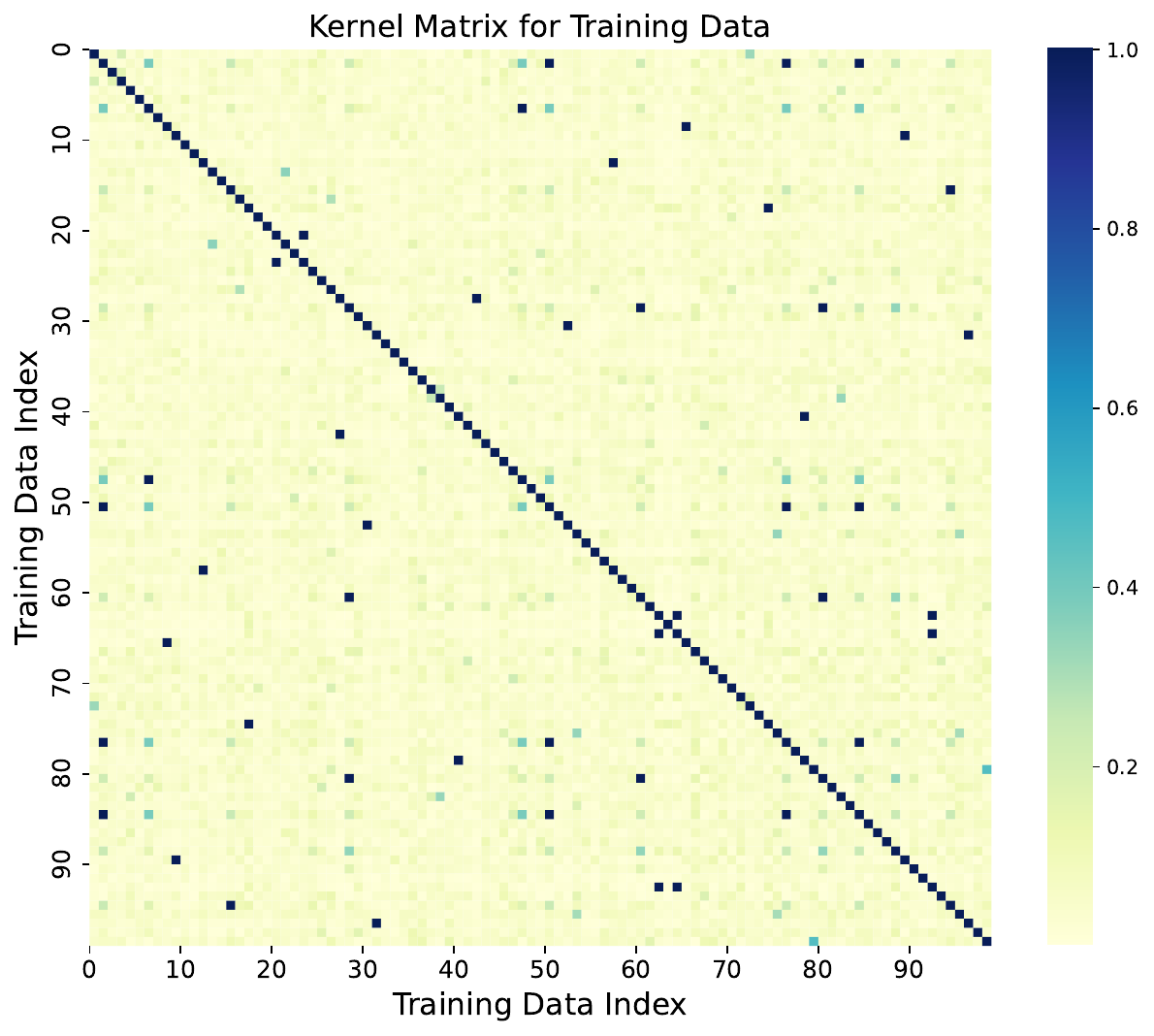}
    \caption{Kernel matrix for QSVM training data using the ZZFeatureMap with full entanglement}
    \label{figure:QSVM_Kernel}
\end{figure}

Upon examining the kernel matrices for both the SVM and QSVM models, distinct patterns emerged, providing insights into the data's behavior in the corresponding feature spaces. In Fig. \ref{figure:SVM_Kernel}, which presents the kernel matrix for SVM, we observe a diagonal with values of 1, as expected, since this diagonal represents the similarity of data points with themselves. More notably, there are numerous high values in the off-diagonal regions, suggesting that many data points appear similar to each other in the feature space defined by the RBF kernel. On the other hand, Fig. \ref{figure:QSVM_Kernel}, displaying the kernel matrix for QSVM, also showcases the anticipated diagonal of 1s but with fewer high values in the off-diagonal. This hints at a distinctive feature space created by the quantum feature map (ZZFeatureMap with full entanglement) where most of the data points are not as similar to each other.

\smallskip

\begin{figure}[ht]
\centering
\includegraphics[width=0.6\textwidth]{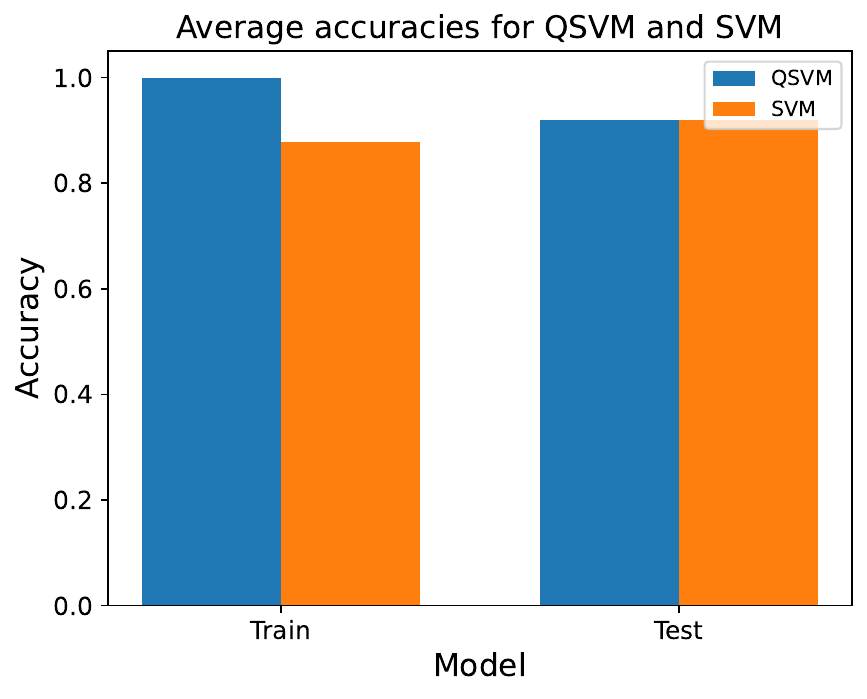}
\caption{Average accuracies for QSVM and SVM}
\label{fig:Average_accuracies}
\end{figure}

In order to fully understand the differences between SVM and QSVM models, a face-to-face accuracy comparison is necessary. Fig. \ref{fig:Average_accuracies} captures this comparative landscape, shedding light on their respective performances across training and testing datasets. When examined, it's apparent that QSVM stands out with perfect accuracy of $100\%$ in the training phase, while the SVM falls behind with an accuracy of $87.89\%$. However, when we transition to the testing dataset, both models exhibit converging performance metrics, showing a closely matched ability to handle unseen data. This near equivalence in test accuracy underscores the robustness inherent to both models, yet the enhanced training performance of QSVM indicates a slight upper hand in grasping and adapting to the specific details and complexities of the dataset.

In our exploration of the SVM and QSVM algorithms, the performance metrics in table \ref{performance_metrics} offer a comprehensive view of how each model performs across multiple evaluation criteria. For the training data, QSVM stands out with perfect scores across all metrics (100\%), reflecting its unparalleled ability to fully capture the nuances of the training set. While SVM produces good results throughout the training phase, doesn't quite match the perfection of QSVM across metrics such as accuracy, precision, sensitivity, specificity, and F1-Score. Interestingly, when evaluating against the test dataset, both models converge towards similar accuracy levels (92\%), demonstrating their robustness in generalizing to unseen data. The observation of perfect train accuracy followed by a small diminished test accuracy suggests that, while the QSVM may have overfit to the training data, it still performs well on unseen data.
Notably, QSVM maintains a competitive advantage in terms of sensitivity (100\%) and F1-Score (93.33\%) for test data, which are crucial metrics in the context of medical diagnostics. These findings, combined, highlight the potential of QSVM in providing a complementary, if not better, approach to traditional SVM,  especially when sensitivity in classification is paramount.

In the cross-validation analysis detailed in Table \ref{cross_validation}, the robustness of the QSVM and the classical SVM is examined through a 10-fold cross-validation method. The QSVM demonstrated a dynamic range in cross-validation performance with a standard deviation of 0.13, and a mean score of 0.83. In contrast, the SVM exhibited commendable stability in cross-validation with a standard deviation of 0.11 and a slightly higher mean score of 0.84. These findings suggest distinct behaviors of the two models when subjected to varying subsets of data, which will be further investigated in the discussion section.

\begin{table}[ht]
\centering
\caption{Performance Metrics (\%)}
\begin{tabular}{|c|c|c|c|c|c|c|}
\hline
\text{ Classifier } & \text{ Class } & \text{ Accuracy } & \text { Precision } & \text{ Sensitivity } & \text{ Specificity } & \text{ F1-Score } \\
\hline
\text{ QSVM } & \text{ Train } & 100 & 100 & 100 & 100 & 100 \\
\cline{2-7}
& \text{ Test } & 92 & 87.5 & 100 & 81.81 & 93.33 \\
\cline{1-7}
\multirow{2}{*}{\text{ SVM }} & \text{ Train } & 87.89 & 89.13 & 85.42 & 90.20 & 87.23 \\ 
\cline{2-7}
& \text{ Test } & 92 & 92.85 & 92.86 & 90.91 & 92.86 \\
\hline
\end{tabular}
\label{performance_metrics}
\end{table}

\begin{table}[ht]
\centering
\caption{Cross-validation results for QSVM and SVM}
\begin{tabular}{|c|c|c|}
\hline
\textbf{Model} & \textbf{QSVM} & \textbf{SVM} \\ \hline
Fold 1 & 0.84615385 & 0.69230769 \\ \hline
Fold 2 & 0.76923077 & 0.76923077 \\ \hline
Fold 3 & 0.53846154 & 0.61538462 \\ \hline
Fold 4 & 0.92307692 & 0.84615385 \\ \hline
Fold 5 & 1.00000000 & 0.91666667 \\ \hline
Fold 6 & 0.91666667 & 0.91666667 \\ \hline
Fold 7 & 0.83333333 & 1.00000000 \\ \hline
Fold 8 & 0.75000000 & 0.91666667 \\ \hline
Fold 9 & 1.00000000 & 0.83333333 \\ \hline
Fold 10 & 0.75000000 & 0.91666667 \\ \hline
\textbf{Mean Score} & \textbf{0.83} & \textbf{0.84} \\ \hline
\textbf{Std. Deviation} & \textbf{0.13} & \textbf{0.11} \\ \hline
\end{tabular}
\label{cross_validation}
\end{table}

\begin{figure}[ht]
    \centering
    \begin{minipage}[b]{0.48\textwidth}
        \centering
        \includegraphics[width=\textwidth]{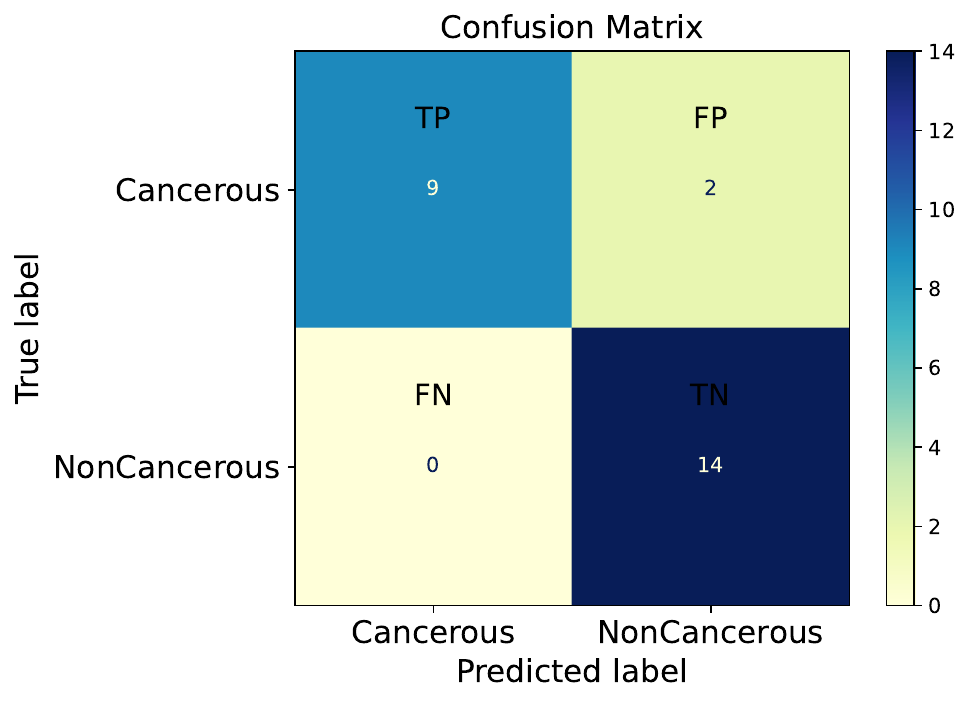}
        \subcaption{}
        \label{fig:qsvm}
    \end{minipage}
    \hfill
    \begin{minipage}[b]{0.48\textwidth}
        \centering
        \includegraphics[width=\textwidth]{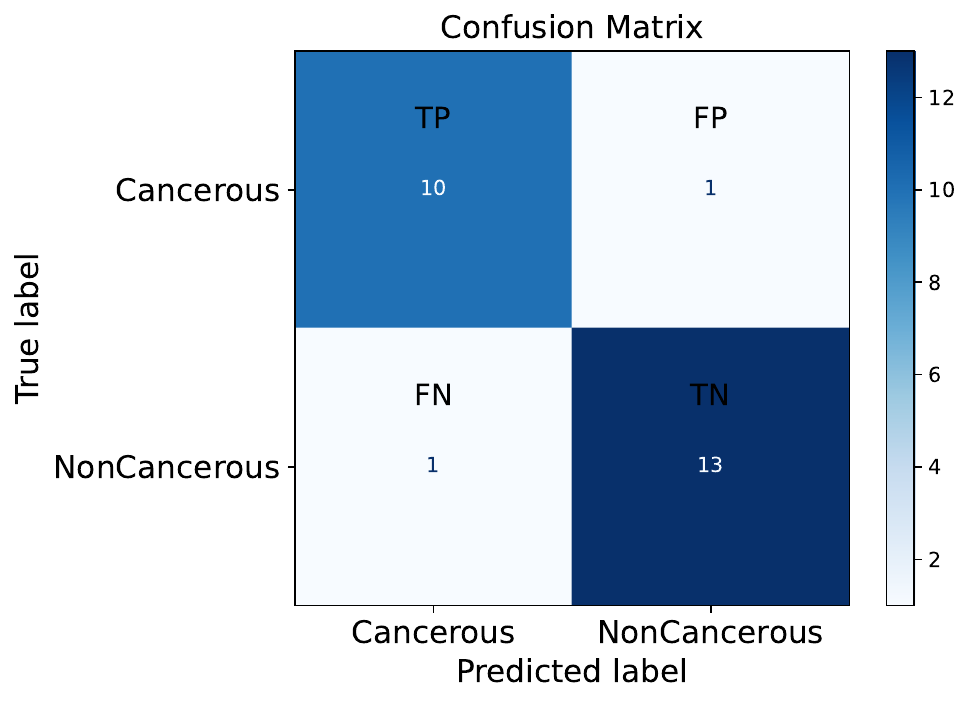}
        \subcaption{}
        \label{fig:svm}
    \end{minipage}
    \caption{Confusion matrices obtained on test data for (a) QSVM and (b) SVM.}
    \label{fig:confusion_matrices}
\end{figure}

Fig. \ref{fig:confusion_matrices} presents the confusion matrices derived from the test data for both QSVM and SVM. Such matrices are pivotal for understanding the predictive behavior of classifiers, providing insights into the correct and incorrect predictions made by the models. For the QSVM in Fig. \ref{fig:qsvm}, a total of 9 instances were accurately predicted as cancerous, denoted as True Positives (TP). Additionally, the model also recorded 14 True Negatives (TN), correctly identifying non-cancerous instances. However, there were 2 instances, termed False Positives (FP), where the QSVM misclassified non-cancerous instances as cancerous. Notably, the QSVM model demonstrated a robust performance by registering no False Negatives (FN), meaning no cancerous instance was misinterpreted as non-cancerous.

Turning our attention to the SVM results presented in Fig. \ref{fig:svm}, this model too showcased a commendable classification capability. It accurately classified 10 instances as cancerous (TP) and correctly identified 13 instances as non-cancerous (TN). Nevertheless, there were minor misclassifications with 1 instance each being classified as a False Positive (FP) and a False Negative (FN).

Comparatively, both models exhibit strong performance, with SVM having a slightly better true positive rate and fewer false positives. However, QSVM achieves a perfect score in identifying non-cancerous cases without any false negatives. These matrices emphasize the strengths and subtle differences in performance between the quantum and classical approaches to SVM, informing potential areas of model improvement and the particular strengths of each method in differentiating between cancerous and non-cancerous cases.

The visual representations and performance metrics consistently suggest that while both SVM and QSVM are robust classifiers, QSVM's unique feature space grants it an advantage in certain key metrics. Additionally, The kernel matrices further substantiated these numerical observations, revealing how each model interacts with the feature space. The implications and potential reasons behind these findings will be further explored in the subsequent discussion section.

\section{Discussion}

We used two different computational paradigms to identify detailed patterns in our dataset: the conventional support vector machine and its quantum version, the quantum support vector machine. These models were chosen based on their ability to capture nonlinear relationships in data using their respective kernel approaches. While the SVM employs the well-known radial basis function to compute the kernel, the QSVM employs a quantum feature map. This quantum feature map is adept at efficiently managing inner products in a higher-dimensional space.

Based on both SVM and QSVM kernel matrices, as illustrated in Figs. \ref{figure:SVM_Kernel} and \ref{figure:QSVM_Kernel} respectively, we get invaluable insights into how each model processes the dataset in its feature space. In Fig. \ref{figure:SVM_Kernel}, the SVM's kernel matrix reveals extensive high similarity values between data points, suggesting a strongly connected feature space under the RBF kernel. This contrasts with Fig. \ref{figure:QSVM_Kernel}, where the QSVM's kernel matrix portrays a more spread-out feature space, characterized by a sparser pattern of high values. This advantage stems from the fact that the ZZ Feature map in QSVM leverages the unique properties of quantum physics, such as superposition and entanglement. Specifically, the full entanglement topology ensures that our quantum states are intrinsically interconnected. This connection allows for a richer exploration of the feature space, aka feature Hilbert space \cite{schuld2019quantum}, enhancing our understanding of complex correlations within the data. This entanglement topology, combined with the inherent superposition of the quantum system, enables the QSVM to encode data in ways that unveil complex patterns and relationships which is computationally challenging for classical systems.

Relating these observations to the performance metrics presented in Table \ref{performance_metrics}, we find the QSVM exhibits perfect scores across all metrics for the training data. The outstanding performance indicates that the quantum feature map's distinctive, spread-out feature space is effective at distinguishing between classes without any overlap. Meanwhile, the SVM, with its high inter-point similarities kernel matrix, still performs well, though not to the same extent as QSVM. The high similarity in its kernel matrix could account for the slightly reduced performance, with some points potentially being challenging to classify due to their close resemblance in the transformed space.

Transitioning our focus to the test data classification, while QSVM algorithm competes with the classical SVM in terms of accuracy, precision, and specificity, the QSVM model outperformed the SVM model in terms of sensitivity (or true positive rate). The QSVM model achieved perfect sensitivities of 100\% for both the training and test sets, while the SVM model had lower values with a train sensitivity of $85.42\%$ and a test sensitivity of $92.86\%$, indicating the QSVM's stronger confidence in correctly identifying all positive instances and made no false negative. The QSVM model's superior sensitivity makes it advantageous in scenarios where correctly identifying positive instances is crucial, such as in medical diagnostics, in our specific case study, this applies to the diagnosis of Prostate cancer disease. The quantum feature mapping and kernel used in the QSVM algorithm contribute to its enhanced sensitivity by allowing for more complex representations and better separation between classes, due to its potential to represent data in higher-dimensional quantum spaces. Also, it is predicted that this efficient distinction of data in the quantum feature space promotes an extended performance for large datasets where classical algorithms start to struggle.

The high sensitivity of the QSVM model is particularly beneficial in situations where avoiding false negatives is critical. For instance, in a medical setting, high sensitivity is important to ensure that no patients with a disease are missed. False negatives can have severe consequences in such cases, as patients may go undiagnosed and untreated. In contrast, false positives can lead to unnecessary treatment or harm, but the cost of a false negative is generally considered to be much higher. Therefore, prioritizing sensitivity over other performance metrics (like accuracy or precision) is crucial in medical diagnoses and other applications where the consequences of missing positive instances are severe.

Additionally, the QSVM model slightly outperformed the SVM model regarding the F1 score. The QSVM achieved an F1 score of $93.33 \%$ on the test data, while the SVM model scored 92.86 \%. The F1-score provides a combined measure of precision and sensitivity and can be used to evaluate overall performance, especially in medical settings. Therefore, the QSVM model demonstrates promising performance in the detection of Prostate cancer.

The cross-validation performance of the QSVM, while showing variability indicative of potential overfitting (as flagged by the perfect training performance shown in Table \ref{performance_metrics}), also indicates the model’s sharp responsiveness to the complexities of the Prostate data set—a quality that could be addressed in future work to enhance model performance. On the other side, the SVM's stable performance across cross-validation folds suggests resilience and not overfitting to the same extent as QSVM. Both models present strengths; the SVM offers reliability across datasets while the QSVM demonstrates a high sensitivity in detection. These characteristics suggest that each model could be beneficial in different scenarios within the domain of cancer detection, depending on the specific requirements for sensitivity and stability. Nevertheless, it is important to recognize that these findings are derived from the simulations carried out, and additional research might be necessary to confirm these conclusions in a more general context.

As far as we are aware, our research is the first to apply QSVM algorithm for categorizing Prostate cancer datasets. This study aims to establish a basis for future Prostate cancer research. Beyond contributing to the growing quantum machine learning applications literature, this research contrasts the QSVM with the conventional SVM— a comparison previously unexplored for Prostate cancer data classification. Our findings underscore that the QSVM algorithm presents an advantage in performance—yet not an alternative in terms of performance—to the conventional SVM technique in terms of sensitivity and overall performance which are crucial for diagnosing Prostate diseases.

\section{Conclusion}

In conclusion, this article has highlighted the potential advantages of employing QSVM in Prostate cancer detection within the field of quantum computing. Our findings reveal that QSVM demonstrates superior performance in predicting prostate cancer, exhibiting a high degree of accuracy that is comparable to the classical SVM method. Notably, QSVM showed an increase of $7.14\%$ in sensitivity, and a high F1-Score compared to the classical SVM approach. These improvements are particularly meaningful in the medical field, as they indicate a high ability of QSVM to correctly identify positive cases, thereby significantly reducing the risk of false negatives. This enhancement combined with the potential speed of QSVM is crucial in early detection and effective treatment of prostate cancer, potentially leading to better patient outcomes and survival rates.

By increasing Precision, QSVM and SVM reduce the likelihood of false positives, thereby lessening patient anxiety and the need for unnecessary follow-up procedures. The improvement in QSVM's sensitivity is particularly significant as it suggests a lower chance of missed diagnoses, an essential factor in strategies that provide early interventions to save lives. Furthermore, higher F1-Score, underscores QSVM's overall effectiveness and reliability as a diagnostic tool.

Moving ahead, this research presents both exciting opportunities and significant challenges for future exploration. A paramount challenge is that we intend to thoroughly investigate the overfitting tendencies observed, particularly within the QSVM model. To tackle this, our future work will be twofold: firstly, we will expand our dataset to include a broader spectrum of data, enhancing the diversity and volume necessary for improved model generalization—a challenge in itself. Secondly, we will refine our model to better capture the nuances of this expanded dataset. In parallel, we will pursue the integration of QSVM in real-time clinical settings, an effort that will not only test the model's robustness but also its translational value in healthcare. Moreover, we see significant opportunities in adapting the QSVM framework for the detection and diagnosis of a wider range of cancers and diseases, potentially revolutionizing multiple areas of medical diagnostics.

The promising results of QSVM in prostate cancer detection represent a significant potential in the intersection of quantum computing and healthcare.  We are optimistic that continued research and development in this field will lead to more robust, efficient, and accessible diagnostic tools, ultimately improving patient care and treatment outcomes in oncology and beyond.

\section*{Declarations}

\subsection*{Ethical Approval}
Not Applicable

\subsection*{Availability of Supporting Data}
The data supporting the findings of this study are available from Kaggle's Prostate Cancer dataset, which is publicly accessible.

\subsection*{Competing Interests}
The authors declare that they have no competing interests.

\subsection*{Funding}
Not Applicable

\subsection*{Authors' Contributions}
This work was carried out in collaboration between all authors. The author conducted the quantum machine learning model experiments and analyzed the data. The co-supervisor and supervisor provided guidance and oversight for the project. All authors read and approved the final manuscript.

\subsection*{Acknowledgments}
Not Applicable

\bibliographystyle{unsrtnat}
\bibliography{sample}

\end{document}